\documentclass{article}

\usepackage{PRIMEarxiv}

\usepackage[utf8]{inputenc} 
\usepackage[T1]{fontenc}    
\usepackage{hyperref}       
\usepackage{url}            
\usepackage{booktabs}       
\usepackage{amsfonts}       
\usepackage{nicefrac}       
\usepackage{microtype}      
\usepackage{lipsum}
\usepackage{fancyhdr}       
\usepackage{graphicx}       
\usepackage{amsmath}
\graphicspath{{media/}}     

\title{Superposition through Active Learning Lens}

\author{ Akanskha Devkar\thanks{\textit{\underline{}}: 
  \textbf{Independent Researcher, Correspondence to: akankshaanc@gmail.com}} 
}

\begin{document}
\maketitle

\begin{abstract}
Superposition or Neuron Polysemanticity are important concepts in the field of interpretability and one might say they are these most intricately beautiful blockers in our path of decoding the Machine Learning black-box. 
The idea behind this paper is to examine whether it is possible to decode Superposition using Active Learning methods. While it seems that Superposition is an attempt to arrange more features in smaller space to better utilize the limited resources, it might be worth inspecting if Superposition is dependent on any other factors. This paper uses CIFAR-10 and Tiny ImageNet image datasets and the ResNet18 model and compares Baseline and Active Learning models and the presence of Superposition in them is inspected across multiple criteria, including t-SNE visualizations, cosine similarity histograms, Silhouette Scores, and Davies-Bouldin Indexes. Contrary to our expectations, the active learning model did not significantly outperform the baseline in terms of feature separation and overall accuracy. This suggests that non-informative sample selection and potential overfitting to uncertain samples may have hindered the active learning model's ability to generalize better suggesting more sophisticated approaches might be needed to decode superposition and potentially reduce it.
\end{abstract}

\section{Introduction}
\textbf{Superposition:}
\\For most, when one hears the term, Superposition it is immediately clubbed with quantum mechanics, the very intriguing quantum entanglement, and always reminds us of the ‘thought experiment’ by Erwin Schrödinger in 1935. But the circuits thread\cite{olah2020zoom}, introduced this word Superposition in a new light, a new meaning for anyone interested in the field of mechanistic interpretability. Just as Schrödinger’s cat can be both alive and dead at the same time in his thought experiment, it was observed how a neuron can activate for both a car wheel and a dog snoot\cite{olah2020zoom}. The authors of Zoom in: An Introduction to Circuits observed that neurons had connections between them which were meaningful algorithms helping them make important and factual decisions, they called these circuits. Upon rigorously studying the circuits, they found a car-detecting neuron at a certain layer but to their surprise, they observed that, in the next layer, the model spreads its car features over several neurons that seem to primarily be detecting dogs. They termed these as ‘Polysemantic’ neurons, and they use the term ‘Superposition’ for the phenomenon of the model having a “pure neuron” and then mixing it up with other features.\cite{olah2020zoom}

In another research work to decode superposition, we saw how and when the models represent more features than they have neurons. Particularly through this work, we saw how when features are sparse, superposition allows compression beyond what a linear model would do, at the cost of "interference" that requires nonlinear filtering. It was also observed that with dense features, the model learns to represent an orthogonal basis of the most important two features, and the lesser important features are not represented. But upon making the features sparse, this changes, not only can models store additional features in superposition by tolerating some interference, but it was seen that, at least in certain limited cases, models can perform computation while in superposition.\cite{elhage2022toy}
In the superposition hypothesis, features are represented as almost orthogonal directions in the vector space of neuron outputs. Since the features are only almost orthogonal, one feature activating looks like other features slightly activating. Tolerating this "noise" or "interference" comes at a cost. But for neural networks with highly sparse features, this cost may be outweighed by the benefit of being able to represent more features.\cite{elhage2022toy}
Is it possible to somehow make a model understand the most important features first and manipulate it to arrange these important features with higher priority?

'How can we control whether superposition and polysemanticity occur?’ was one of the open questions mentioned by the authors of ‘Toy Models of Superposition’ and this paper explains an attempt to do so with the help of Active Learning.

\begin{figure*}
    \includegraphics[width=\textwidth]{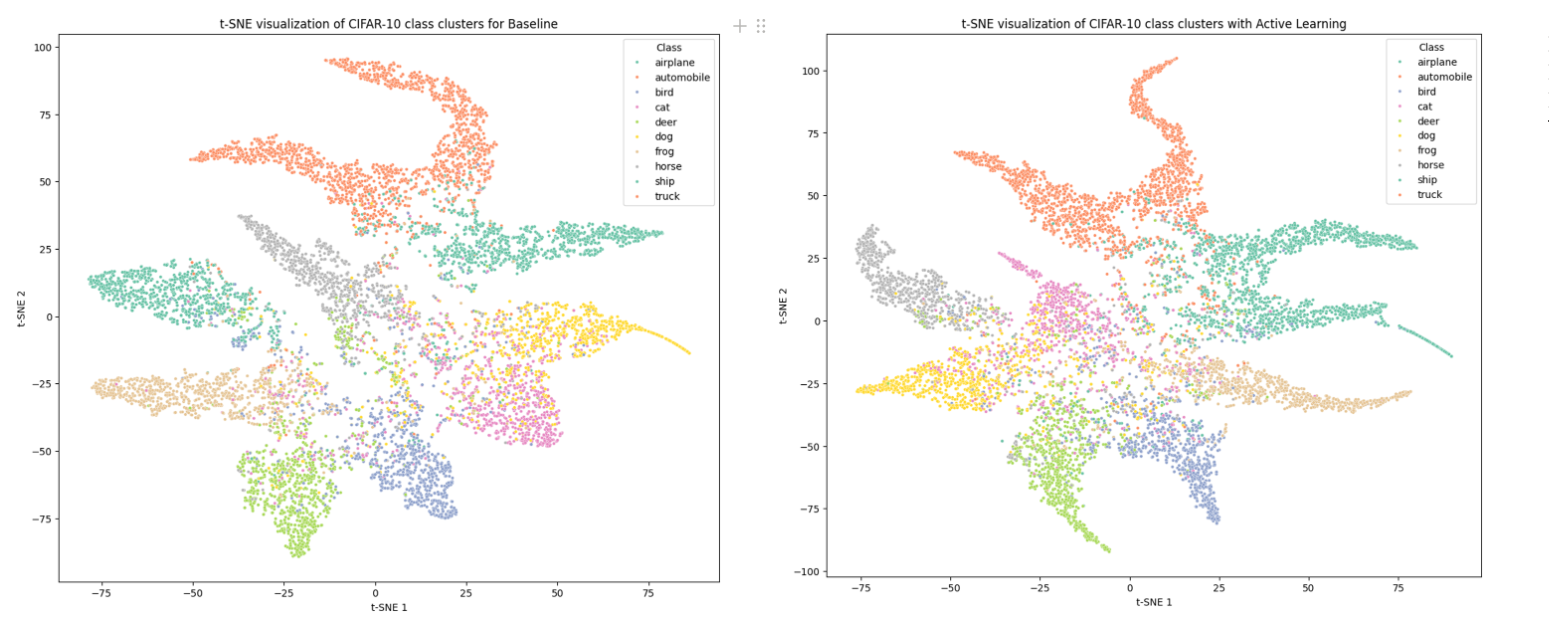}
    \caption{t-SNE visualization of class clusters for CIFAR-10 dataset for Baseline and Active Learning}
\end{figure*}

\begin{figure*}
    \includegraphics[width=\textwidth]{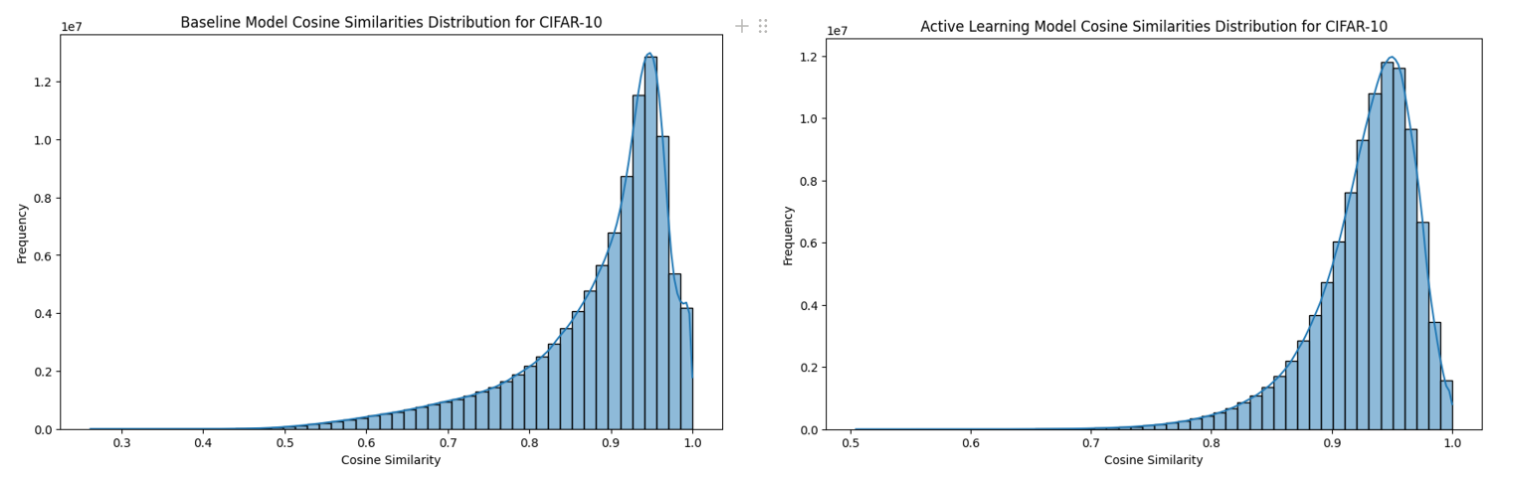}
    \caption{Cosine Similarity Histograms for CIFAR-10 dataset for Baseline and Active Learning}
\end{figure*}

\textbf{Active Learning:}
\\If we assume visual recognition to be an equation, it will most likely depend upon the three variables which are the features, the classifiers, and the data selected, we will see that active learning is all about the data variable of this equation.

Although Convolutional Neural Networks (CNNs) are universally successful in many tasks, they have a major drawback; they need a very large amount of labeled data to be able to learn their large number of parameters. More importantly, it is almost always better to have more data since the accuracy of CNNs is often not saturated with increasing dataset size. Hence, there is a constant desire to collect more and more data.\cite{sener2017active}

The goal of active learning is to find effective ways to choose data points to label, from a pool of unlabeled data points, to maximize the accuracy.\cite{sener2017active}

Active learning is a branch of machine learning where a learning algorithm can actively query a human user to learn on unsupervised data or in some supervised data, it can be used to select the data it wants to learn from.
There are various methods to efficiently train deep networks on large datasets, suggesting that proper data selection and preprocessing are crucial for optimal performance\cite{krizhevsky2009learning}, active learning is one such method.

Although it is not possible to obtain a universally good active learning strategy\cite{NIPS2004_c61fbef6}, it is observed that active learning is successful in some cases as it achieves a significant reduction in the training required,
along with efficient scaling to a large number of categories and huge data sizes, and achieves state-of-the-art performance as seen in the case of a custom method being able to evaluate the “informativeness” of a sample across multiple classes, making the selection more accurate\cite{Yang2015MultiClassAL}.

Out of the many methods of active learning, uncertainty-based sampling is one method where the sample points for which the model is least certain as to what the correct output should be, are selected for training. This method also shows some promising results as seen from the results in\cite{li2006confidence}, \cite{settles2004active}. From the previous research works it can be assumed that the uncertain samples lie near the decision boundary or the regions of feature space that lack sufficient training data, so by allowing the model to learn more from such sample points by picking them with priority for training should give us some interesting insights into its impact on superposition in the feature space. Also, by not selecting the more confident sample points for training, the model would treat them as redundant and thus result in less occupancy in the overall feature space for them. 
Intuitively, both these factors should have a positive impact on superposition, i.e. reduce it, hence, uncertainty-based sampling was chosen to experiment with. But the results obtained were not in coherence with the intuition at all!

\section{Methodology}
The key idea of this paper is to compare the superposition observed in a model that is trained in the usual way with a model that is trained using the active learning method. Let us look into the building blocks of this experimental study and then understand the experimental procedure in detail.

\textbf{Dataset Description:}
\\Taking inspiration from the work of the Circuits thread\cite{olah2020zoom}, but due to the lack of computational resources, the smaller vision datasets CIFAR-10 and Tiny ImageNet were used for the sake of this experiment. 
CIFAR-10 is a standard benchmark in computer vision, widely used for evaluating and comparing models and algorithms, it comprises 50,000 training images of size 32x32 pixels, divided equally into 10 classes.
The small size of the images and the dataset itself allows for rapid experimentation and iteration, which is crucial for active learning experiments. The smaller size and lower quality of the CIFAR-10 images and their potential impact on superposition were also an interesting point for choosing this dataset along with the lower computational ask. 
The other dataset used in this study is Tiny ImageNet, which is a subset of the ImageNet dataset, resized and reduced in scope to make it more manageable for research tasks. It comprises 100,000 images of size 64x64 pixels, divided equally into 200 classes.
Tiny ImageNet was used as it provides a richer, more complex, and diverse set of images compared to CIFAR-10, making it a natural step up for this experimental study to decode superposition.

\textbf{Model Description:}
\\For this study, ResNet-18 model is used which is a deep CNN with 18 layers. ResNet-18 model’s architecture consists of a combination of convolutional layers, batch normalization, ReLU activations, and residual blocks. It is widely used for image classification tasks due to its balance of depth and computational efficiency.

\begin{figure*}
    \includegraphics[width=\textwidth]{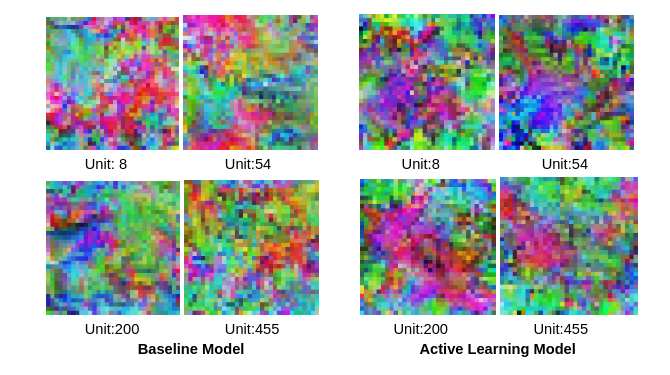}
    \caption{Activation Maximization Visualizations for Units 8,54,200,455 of Layer4.1.conv2 for CIFAR-10 Dataset}
\end{figure*}

\textbf{Training Details:}
\\The baseline model was trained on both the CIFAR-10 dataset and the Tiny ImageNet dataset using a pre-trained ResNet-18 architecture. The final fully connected layer was modified to output 10 classes and 200 classes for the CIFAR-10 dataset and the Tiny ImageNet dataset respectively.
The baseline model was trained for 20 epochs with mixed precision to enhance computational efficiency, and Cross-Entropy Loss was used for calculating the loss.

For the active learning model, the model is trained on both the CIFAR-10 and the Tiny ImageNet dataset in small batches. Active learning is a process of prioritizing the data samples to be trained. This prioritization can be performed using various strategies such as:

1. Uncertainty based
\\- Least Confidence based
\\- Margin of confidence based
\\- Ratio based
\\- Entropy-based
\\2. Streaming based
\\3. Pooling based
\\4. Query by Committee based

For the sake of the experimental study carried out in this paper, least confidence-based uncertainty sampling was used as the sampling strategy where the uncertainty is calculated using (1).

\begin{align}
s_{LC} &= \arg\max_{x} (1 - P(\hat{y}|x)) 
\end{align}

For the CIFAR-10 dataset which has a total of 50,000 training images, the active learning model was trained for 10 loops, on 5000 images at a time in a loop. The first loop was trained on randomly sampled 5000 images out of all the 50,000 training set images. From the inference data obtained on the remaining 45,000 training images, the least confident 5000 images were extracted to form the next training batch for the next loop. This goes on for 10 loops i.e. till the training data is completely exhausted.
For the Tiny ImageNet dataset consisting of 100,000 training images, the active learning model was trained for 10 loops, on 10,000 images at a time per loop. The sampling was similar to the way earlier described, random for the first loop and then using Least confident samples from the inference data.

\section{Visualizations and Statistical Insights:}
For analyzing the impact on superposition and comparing this impact between the Baseline model and the Active Learning model, the following methods are used:
\begin{itemize}
\item \textbf{Visual Inspection of the activation maximization obtained using the Lucent Library}:
If the models are trained on high-quality image data, one can observe the activation maximization for each neuron and visually inspect the amount of superposition in the last layers, and compare two models accordingly as seen in\cite{openai2020microscope}.
The images in \cite{openai2020microscope} were obtained using Lucid Library, for this paper, Lucent Library (Lucid for Pytorch) has been used to get similar insights\cite{greentfrapp2020lucent}.
\item  \textbf{t-SNE plots to visualize Class Clusters}: By plotting the feature vectors using t-SNE\cite{mu2021compositional}, it is possible to visually inspect how well the features of different classes are separated or if they overlap significantly, indicating potential superposition. These plots provide an insight into how distinct or entangled the feature vectors are which in turn helps us guess the amount of superposition.
\item  \textbf{Cosine Similarity Histograms}: These plots show the cosine similarity between the representations learned by the different neurons in the model.
\item  \textbf{Cosine Similarity Statistics}: These help in quantifying the average similarity and the spread of similarities. For example, when the mean is high and the standard deviation is low, it indicates a higher superposition, while a broader spread would suggest more distinct features, so less superposition.
\item  \textbf{Silhouette Score and Davies-Bouldin Index}: These two values are robust metrics to compare clustering quality across different models.
\end{itemize}

\begin{figure*}
    \includegraphics[width=\textwidth]{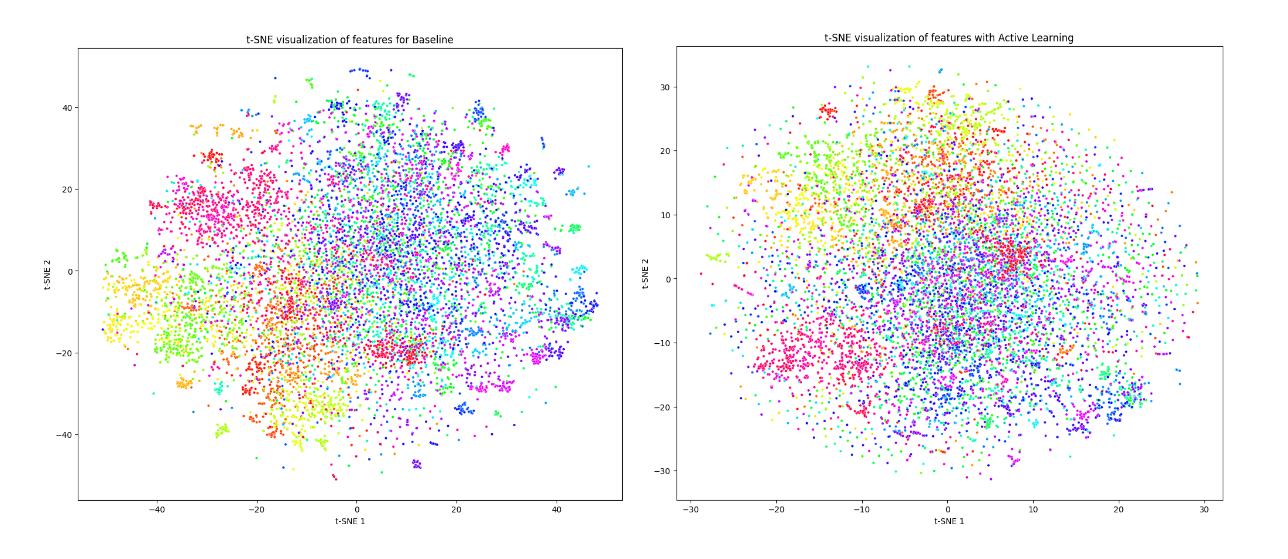}
    \caption{t-SNE visualization of class clusters for Tiny ImageNet dataset for Baseline and Active Learning}
\end{figure*}

\begin{figure*}
    \includegraphics[width=\textwidth]{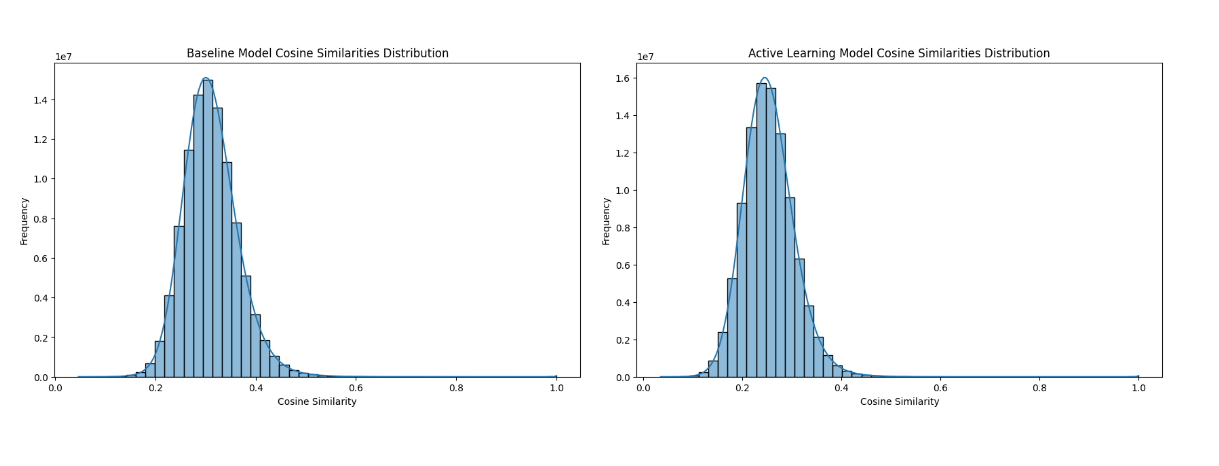}
    \caption{Cosine Similarity Histograms for Tiny ImageNet dataset for Baseline and Active Learning}
\end{figure*}

\section{Experimental Results}
\textbf{For CIFAR-10 Dataset:}
\\ The t-SNE visualizations for CIFAR-10 as seen in Figure 1 reveal that the baseline model exhibits distinct and well-separated clusters for different CIFAR-10 classes, indicating less superposition. Whereas, the active learning model shows more overlap between clusters, and as we can see the cat class cluster is the weakest and seems to be scattered across other classes. Interestingly, the cat class was the most picked class during the active learning loops, implying that while active learning focuses on uncertain samples, it causes less distinct feature representations, leading to more superposition in the feature space. 
From the Cosine Similarity Histograms for CIFAR-10 observed in Figure 2, we can see that the distribution for both Baseline and Active Learning models are mostly similar. But the Active Learning model has a narrower spread compared to the Baseline indicating fewer lower similarity values and it also shows a tightly concentrated distribution around higher similarity values. This suggests a more uniform feature distribution in the case of Active Learning.
This insight is also corroborated by the Cosine similarity statistics as the active learning model has a higher mean cosine similarity of 0.927 compared to the Baseline at 0.885 indicating features in Active Learning are more similar to each other.
But the Baseline model has a higher silhouette score of 0.212 compared to the active learning model of 0.189, so better-defined clusters, suggesting that the baseline model might have better class separability in the feature space.

The Davies-Bouldin index for both models is quite similar, with the Active Learning model having a slightly lower index of 1.542 compared to the Baseline model of 1.551.
The higher mean cosine similarity, lower standard deviation, and lower silhouette score collectively suggest that the active learning model has more superposition. The feature vectors are more similar and less distinct, leading to overlapping features among different classes.
And finally, we can also compare the activation maximization for a few randomly selected neurons in the final convolutional layer 4.1.conv2 of the model for both Baseline and Active Learning, in Figure 3.
As the CIFAR-10 image data size is very small, it is hard to interpret what features the neurons are exactly activating for, but the overall more noisy patterns throughout the Active Learning Units indicate more superposition. 

\begin{figure*}
    \includegraphics[width=\textwidth]{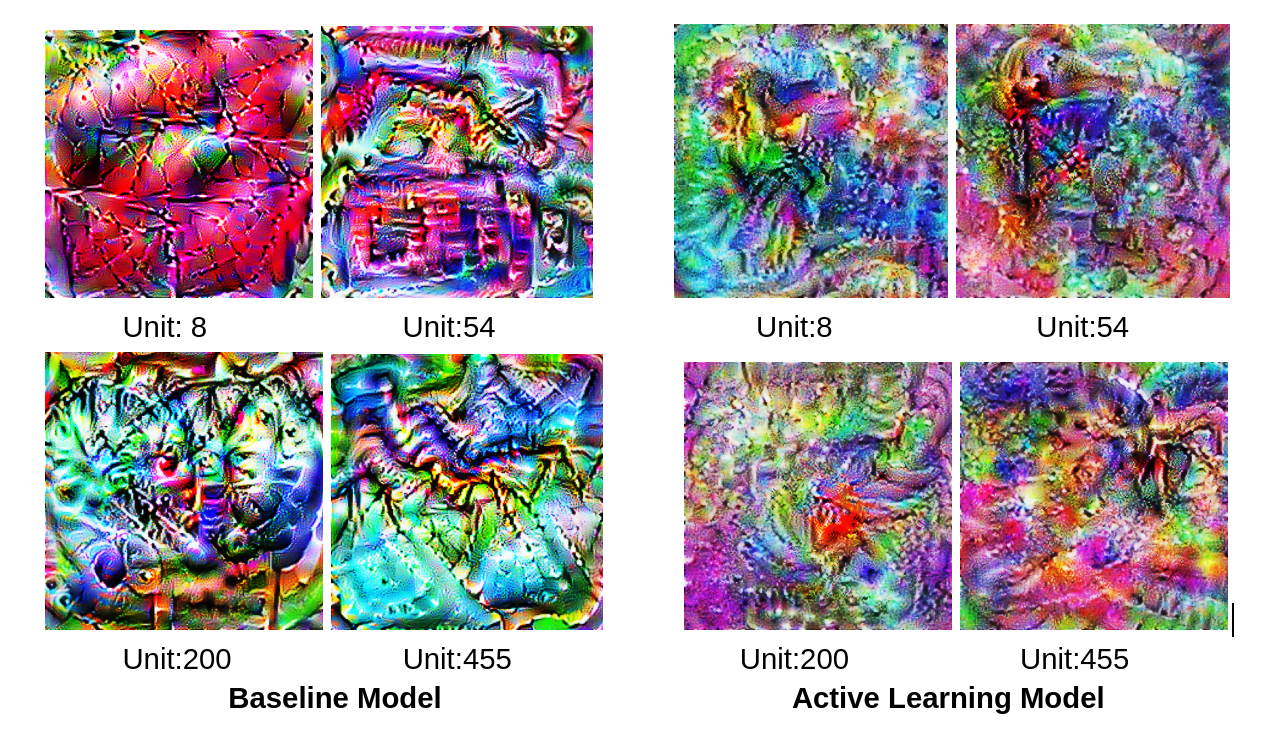}
    \caption{Activation Maximization Visualizations for Units 8,54,200,455 of Layer4.1.conv2 for Tiny ImageNet Dataset}
\end{figure*}

\textbf{For Tiny ImageNet Dataset:}
\\Figure 4 shows t-SNE visualizations for Tiny ImageNet, and it shows a somewhat better class cluster separation for Baseline, whereas the clustering in Active Learning seems to have no strict boundaries, indicating a lot of overlap and in turn more superposition. 
Figure 5 shows Cosine Similarity Histograms for Tiny ImageNet, both Baseline and Active Learning distributions are quite similar in terms of mean, spread, and shape. The slight difference in spread might indicate that the active learning model has slightly more consistent feature representations, but this difference is not substantial.
The Active Learning model has a lower mean cosine similarity 0.254 and a slightly tighter distribution compared to the Baseline at 0.310. This suggests that while the feature vectors are less similar to each other on average, their similarities are more consistent in the active learning model. The Active Learning model has a lower silhouette score of -0.399, compared to the Baseline at -0.348 indicating worse clustering quality. The negative values for both indicate many sample points being assigned to the wrong clusters. The Davies-Bouldin Index of Baseline is much lower than the Active Learning model, indicating higher similarity between clusters in the Active Learning model and suggesting that the clusters are less distinct, which can be corroborated by the t-SNE plot in Figure 4.

Figure 6 shows the activation maximization for randomly selected units from the Layer4.1.Conv2 shows a generally more crowded pattern for the Active Learning model as compared to Baseline which might indicate more superposition.
It should be noted that these units are not reacting to the same features as they are from different models, so they are not ‘seeing’ the same details in the training images so we can not directly compare them. However, due to this task of getting activation maximization being computationally exhaustive, random units were selected and visualized to inspect the level of activity in them.

\section{Conclusion}
For both datasets, Active Learning clearly performs worse overall and seems to have more Superposition which can be endorsed by the different methods of inspection. Interestingly, the Active Learning model in certain cases seems to have more uniform distribution but that does not reflect in its ability to better distinguish between classes. The higher overlap in the class clustering and cosine similarity statistics indicates higher superposition in the Active Learning model.
The Active Learning model's decision boundaries might not have improved significantly as compared to the Baseline when trained on the least confident samples. Instead of creating clearer separations, the  Active Learning model seems to have reinforced existing overlaps or created new ones. It is evident from a lot of related work on Active Learning that it generally tends to boost performance but has failed to do so in the context of this paper’s experiment. A thorough walkthrough of the Active Learning Model’s data picking and the subsequent impact on the feature representations might help understand what exactly caused the higher superposition.

\section{Future Work}
It might be helpful to examine different data sampling strategies from the Active Learning methods to see if there is a difference in the Superposition based on them. Working with a deeper model and a better quality dataset to conduct the same experiment might give a better chance of evaluating the Superposition phenomenon through an Active Learning Lens and subsequently decoding if Superposition is in any way related to the complexity spectrum of features it learns and in turn their packing.

\bibliographystyle{unsrt}

\end{document}